\documentclass[conference]{IEEEtran}
\IEEEoverridecommandlockouts
\usepackage{amsmath,amssymb,amsfonts}
\usepackage{graphicx}
\usepackage{textcomp}
\usepackage{hyperref}
\usepackage{booktabs}
\usepackage{xcolor}
\usepackage[
maxbibnames=999,
]{biblatex}
\usepackage{listings}
\usepackage{makecell}
\def\BibTeX{{\rm B\kern-.05em{\sc i\kern-.025em b}\kern-.08em
    T\kern-.1667em\lower.7ex\hbox{E}\kern-.125emX}}
\usepackage{changes}
\usepackage{listings}
\usepackage{algorithm}
\usepackage{algpseudocode}
\makeatletter
\newcommand{\linebreakand}{%
  \end{@IEEEauthorhalign}
  \hfill\mbox{}\newline
  \mbox{}\hfill\begin{@IEEEauthorhalign}
}
\makeatother
    
\begin{document}

\title{PillagerBench: Benchmarking LLM-Based Agents in Competitive Minecraft Team Environments}

\author{%
  \IEEEauthorblockN{%
    Olivier Schipper\textsuperscript{1} \quad
    Yudi Zhang\textsuperscript{1} \quad
    Yali Du\textsuperscript{2} \quad
    Mykola Pechenizkiy\textsuperscript{1} \quad
    Meng Fang\textsuperscript{3,1}%
  }%
  \\
  \vspace{1ex} 
  \IEEEauthorblockA{%
    \textsuperscript{1}Department of Mathematics and Computer Science, Eindhoven University of Technology, Eindhoven, Netherlands\\
    \textsuperscript{2}Department of Informatics, King’s College London, London, UK\\
    \textsuperscript{3}Department of Computer Science, University of Liverpool, Liverpool, UK\\
    \{o.t.schipper, y.zhang5, m.pechenizkiy\}@tue.nl, yali.du@kcl.ac.uk, Meng.Fang@liverpool.ac.uk
  }%
}
\maketitle
\begin{abstract}
LLM-based agents have shown promise in various cooperative and strategic reasoning tasks, but their effectiveness in competitive multi-agent environments remains underexplored. To address this gap, we introduce PillagerBench, a novel framework for evaluating multi-agent systems in real-time competitive team-vs-team scenarios in Minecraft. It provides an extensible API, multi-round testing, and rule-based built-in opponents for fair, reproducible comparisons. We also propose TactiCrafter, an LLM-based multi-agent system that facilitates teamwork through human-readable tactics, learns causal dependencies, and adapts to opponent strategies. Our evaluation demonstrates that TactiCrafter outperforms baseline approaches and showcases adaptive learning through self-play. Additionally, we analyze its learning process and strategic evolution over multiple game episodes. To encourage further research, we have open-sourced PillagerBench, fostering advancements in multi-agent AI for competitive environments.
\end{abstract}

\section{Introduction}

Witnessing rapid advancements, Large Language Models (LLMs) have emerged as powerful tools for complex reasoning, decision-making, and facilitating multi-agent collaboration~\cite{zhao2023survey, Wang_2024, xi2023risepotentiallargelanguage}. This has driven increasing interest in developing cooperative multi-agent systems~\cite{chen2023agentversefacilitatingmultiagentcollaboration,hong2024metagptmetaprogrammingmultiagent}, leading to the creation of benchmarks based on diverse cooperative games such as Minecraft~\cite{dong2024villageragent} and Overcooked~\cite{agashe2023llm}. Minecraft, in particular, has become an important platform due to its open-ended environment and rich state and action spaces~\cite{wang2023voyageropenendedembodiedagent, yu2024adamembodiedcausalagent}.

However, current Minecraft-based benchmarks mainly address cooperative tasks characterized by stationary dynamics and fixed objectives, making them insufficient for evaluating adaptability and strategic decision-making in competitive, dynamic environments. Traditional reinforcement learning benchmarks like StarCraft Multi-Agent Challenge (SMAC) \cite{samvelyan19smac} and Lux AI Challenge \cite{lux-ai-season-2} introduce instability and nonstationarity through competitive adversaries but lack the rich, open-ended interactions found in Minecraft. Bridging this gap by integrating both cooperative and competitive elements within a single dynamic environment is essential to rigorously assess the adaptability and generalizability of advanced multi-agent systems.

\begin{figure}
    \centering
    \includegraphics[width=1\linewidth]{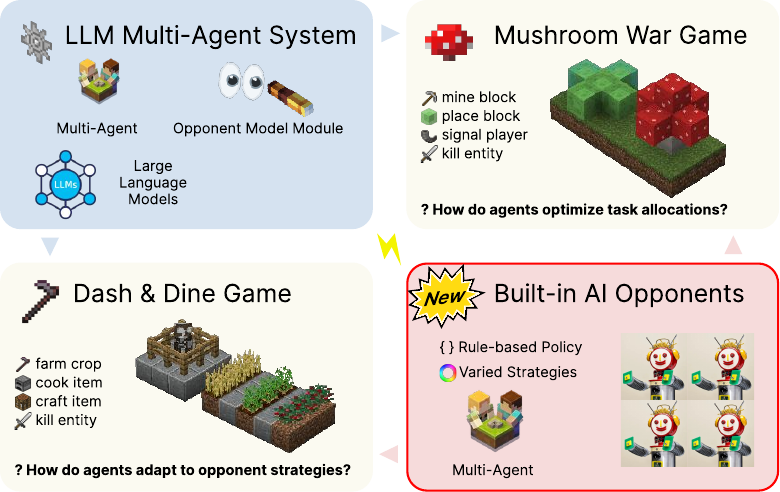}
    \caption{PillagerBench introduces built-in AI opponents, making it the first multi-scenario benchmark designed to evaluate the \textbf{competitive} capabilities of multi-agent systems within the real-world context of Minecraft.}
    \label{fig:pillagerbench}
\end{figure}

To address this gap, we introduce \textbf{PillagerBench}, the first benchmark which is designed to evaluate LLMs in competitive, team-vs-team multi-agent systems, based on the real-time complex environment of Minecraft. Unlike previous work, PillagerBench features multiple distinct competitive scenarios that require agents to balance \textit{cooperation} with teammates and \textit{competition} against opponents in a dynamic and resource-constrained environment.
PillagerBench contains two competitive game scenarios: \textcolor{teal}{\emph{Mushroom War}} and \textcolor{teal}{\emph{Dash \& Dine}}, with each evaluating different capacity of LLMs. {\emph{Mushroom War}} requires agents to efficiently allocate tasks and perform their action under time constraints.  {\emph{Dash \& Dine}} challenges agents to plan ahead, adapt to opponent strategies, and manage complex causal dependencies in a dynamic environment with numerous interaction possibilities. By offering adversarial scenarios, PillagerBench enables more comprehensive evaluation of multi-agent adaptability for their peers and strategy generalization to various opponent states.

Facing the challenges in PillagerBench, we introduce \textbf{TactiCrafter}, a novel LLM-based multi-agent system which excels in task allocation and causality information usage. TactiCrafter features a Tactics Module that generates high-level strategies and assigns sub-tasks to agents, a Causal Model that aggregates causal knowledge from gameplay, and an Opponent Model that infers enemy strategies based on observations (Figure~\ref{fig:tacticrafter}). These components enable agents to learn, adapt, and overcome through repeated play.
Evaluating the proposed method TactiCrafter in PillagerBench, we demonstrate that TactiCrafter outperforms baseline approaches, including random strategies and LLM-based Chain-of-Thought (CoT) reasoning \cite{wei2023chainofthoughtpromptingelicitsreasoning}. Furthermore, our results show that TactiCrafter can adapt to adversarial agents and learn from repeated self-play. Through ablation studies, we analyze the impact of different LLM configurations and modules, providing insights into optimal setups for competitive multi-agent decision-making.

To encourage further research and ensure reproducibility, we make PillagerBench and our code publicly available.\footnote{\href{https://github.com/aialt/PillagerBench}{https://github.com/aialt/PillagerBench}}

\section{Background}
In this section, we briefly introduce the preliminaries and then explore the work related to this research.
\subsection{Preliminaries}

To ensure the accessibility of this paper to a broad audience of computer science students, we provide a brief overview of relevant background concepts.

\paragraph{Game Theory and Zero-Sum Games}  
Game theory~\cite{owen2013game} provides a framework for analyzing strategic interactions, and in this paper, we model the team-vs-team scenario as a \textit{zero-sum}, \textit{simultaneous}, and \textit{stochastic} game with \textit{incomplete information}. Since the utility function is the difference in points between the two teams, the game is inherently zero-sum: $U_A = P_A - P_B$ and $U_B = P_B - P_A$. Each team selects an optimal strategy $s_A \in S_A$ or $s_B \in S_B$ to maximize its expected utility, i.e., $\max_{s_A} \mathbb{E} [ U_A ]$ and $\max_{s_B} \mathbb{E} [ U_B ]$. The interaction between both teams leads to a Nash equilibrium where neither can improve unilaterally, formulated as $\max_{s_A} \min_{s_B} U_A(s_A, s_B) = \min_{s_B} \max_{s_A} U_A(s_A, s_B)$. This minimax principle guides strategic decision-making, and since teams must adapt to an opponent’s evolving strategies, adversarial learning techniques like self-play and opponent modeling are essential for improving agent performance in this competitive setting.

\paragraph{Causal Graphical Models}
A Causal Graphical Model captures the causal relationships within a system~\cite{peters2017elements} by representing how variables $\{X_1,\dots,X_n\}$ influence one another. It consists of a probability distribution over these variables and is structured as a directed acyclic graph (DAG), known as a causal graph. In this graph, nodes correspond to individual variables, while directed edges indicate direct causal effects, signifying that changes in one variable ($X_i$) directly impact another ($X_j$). This framework provides a structured approach to understanding causality within complex systems.

\subsection{Related Work}

\textbf{Minecraft Agents.}
Minecraft, a sandbox video game, provides a versatile platform for AI research due to its open-ended nature and rich set of possible actions and interactions \cite{wang2023voyageropenendedembodiedagent,nottingham2023embodiedagentsdreampixelated,yu2024adamembodiedcausalagent,zhu2023ghostminecraftgenerallycapable}. The game allows for complex scenarios involving resource management, spatial reasoning, and interactions with other entities. 
Approaches built on MineRL~\cite{guss2019minerllargescaledatasetminecraft} take video as input and perform fine-grained actions. OpenAI's VPT~\cite{baker2022videopretrainingvptlearning} uses Reinforcement Learning with large amounts of pre-training on video. DECKARD~\cite{nottingham2023embodiedagentsdreampixelated} further improves this by adding knowledge from LLMs to assist the reinforcement learning exploration.

Text-based approaches require a translation layer between the 3D world of Minecraft and pure text. Mineflayer~\cite{mineflayer} provides a JavaScript interface for Minecraft where agents can interact with Minecraft through JavaScript commands. For example taking observations in the form of JSON output from a command, or performing actions by writing JavaScript code that uses the commands.
Notable works that use Mineflayer are Voyager~\cite{wang2023voyageropenendedembodiedagent} which demonstrates the potential of LLMs as generally capable agents in Minecraft with open-ended exploration.
Adam~\cite{yu2024adamembodiedcausalagent} improves on Voyager~\cite{wang2023voyageropenendedembodiedagent} by adding a causal module that constructs a causal graph of the tech tree using intervention-based causal discovery and a perception model that generates descriptions of what the bot is seeing. 
Interventions during sampling~\cite{eberhardt2007interventions} allow us to analyze the causal dependencies between variables, which serves as the gold standard for causal discovery~\cite{spirtes2001causation, peters2017elements}.

\paragraph{Multi-Agent Minecraft Environments}
In the multi-agent paradigm, \cite{yocum2023mitigating} facilitates cooperation between multiple Voyager-style agents with diverging goals by having them negotiate a mutually beneficial contract.
VillagerAgent~\cite{dong2024villageragent} introduces the VillagerBench multi-agent benchmark for complex tasks in Minecraft, and a top-down orchestrated multi-agent system called VillagerAgent specialized in solving complex sub-task dependencies.
MineLand~\cite{yu2024minelandsimulatinglargescalemultiagent} offers a large-scale multi-agent simulation with multimodal senses and physical needs. 

\paragraph{Competitive Multi-Agent Benchmarks}
Team-vs-team scenarios serve as an excellent testbed for evaluating multi-agent collaboration. These scenarios require multi-agent systems to balance cooperation with competition, adapt to the strategies of opponents, and optimize their performance under dynamic conditions. Real-world parallels include team sports, military operations, and collaborative robotics, where success often hinges on effective teamwork and strategic adaptation.
Outside of Minecraft there are a few benchmarks for multi-agent team-vs-team scenarios like the StarCraft Multi-Agent Challenge (SMAC)~\cite{samvelyan19smac} and Lux AI Challenge~\cite{lux-ai-season-2}. The challenge provided in these benchmarks is similar to a scenario in PillagerBench, in that they are both team-vs-team scenarios with limited time. However they are single scenarios designed for specialized agents, thus not capable of testing the generalization ability of the multi-agent system.
FightLadder~\cite{li2024fightladderbenchmarkcompetitivemultiagent} tests competitive reinforcement learning agents in multiple games against various opponents, but is limited to one versus one scenarios.

This project builds on these foundations by introducing a new benchmark in Minecraft, focusing on team dynamics, adversarial strategic interactions, and generalization ability.

\section{Benchmark - PillagerBench}

PillagerBench offers competitive team-vs-team scenarios in Minecraft (Figure~\ref{fig:pillagerbench}) via a flexible interface that supports any number of teams and agents, using either rule-based or LLM-based systems. 
In the following section, we detail the benchmark's architecture, scenarios, and built-in opponents.

\subsection{Environment Design}
\begin{figure}
    \centering
    \includegraphics[width=1\linewidth]{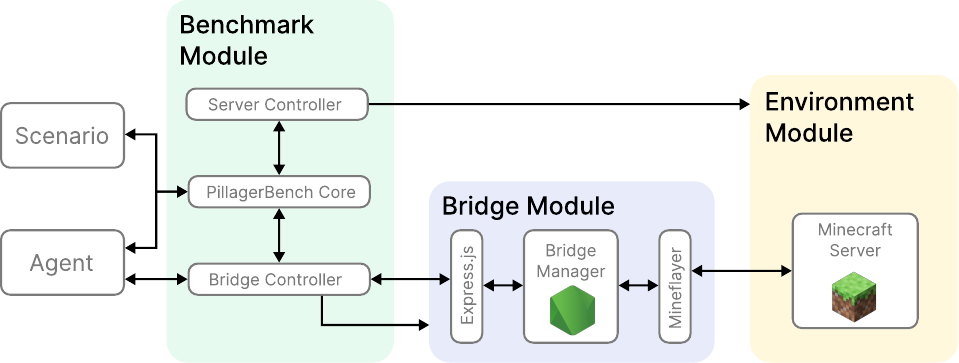}
    \caption{Illustration of the architecture of PillagerBench.}
    \label{fig:pillagerbench_architecture}
\end{figure}

\textbf{Benchmark Components.} PillagerBench features a modular architecture with three main components—the Benchmark, Bridge, and Environment Modules (Figure~\ref{fig:pillagerbench_architecture})—where the Benchmark Module spawns the Minecraft server and one Mineflayer process per agent for scalable performance on consumer hardware~\cite{yu2024minelandsimulatinglargescalemultiagent}. Reliability is ensured via Docker~\cite{merkel2014docker}, which provides a consistent operating system and port availability, while reproducibility is achieved using YAML configurations managed by Hydra~\cite{Yadan2019Hydra} that define all experiment parameters, including scenario lists, episode counts, and agent policies. Each run is logged to a unique folder for later visualization. Moreover, all multi-agent systems in PillagerBench implement a common API that supports three phases—pre-game (access to scenario metadata, team ID, and initial state), game (unlimited actions and observations via Mineflayer instances), and post-game—with resources deliberately limited to ensure fair play. The multi-agent system object persists between episodes, enabling continuous learning.

\textbf{Teams.}
One team is controlled by one multi-agent system. In the general case, a team can have any number of agents and there can be any number of teams.  By default, it includes two scenarios where
two teams of two agents compete in consecutive 2-minute
episodes.  Our benchmark comes with two pre-implemented scenarios, both support two teams—the red team with agents ``Ryn'' and ``Raze'' and the blue team with agents ``Byte'' and ``Blink''—each comprising two agents.

\textbf{Built-in opponents.} PillagerBench provides multiple choice of the opponent team in the game, allowing users to compare their multi-agent systems based on how they perform against the built-in opponents.
The built-in opponents are specialized multi-agent systems for each scenario. They load the JavaScript script with the Mineflayer API for each agent in their team, replace some variables with constants from the scenario, and then execute it during the game phase.  
There is also the \textit{do\_nothing} built-in opponent whose only action is to wait and do nothing. 

\textbf{State Space.}
PillagerBench provides scenario metadata and a sequence of events triggered during action execution. Each event is either a chat (in-game messages or environment feedback) or an observe event (generated after each action), containing detailed information such as chat content and sender, nearby blocks, contents of interacted chests, nearby mobs with distances, agent inventory, and self-status (health, hunger, position, velocity, direction, equipment, environmental context, biome, time, inventory usage count, and elapsed time).

\textbf{Action Space.}
The action space consists of high-level actions, such as mining blocks and slaying mobs. The actions are implemented in the form of JavaScript code, following Voyager~\cite{wang2023voyageropenendedembodiedagent}.
The action code can make use of Mineflayer API or any available control primitives~\ref{table:control_primitives}.
These control primitives are JavaScript functions which perform high-level actions and might have some have additional arguments to customize the behavior. Agents can simply choose to include these control primitives in their executable action code. 

\subsection{Mushroom War} 
\begin{figure}
    \centering
    \includegraphics[width=0.75\linewidth]{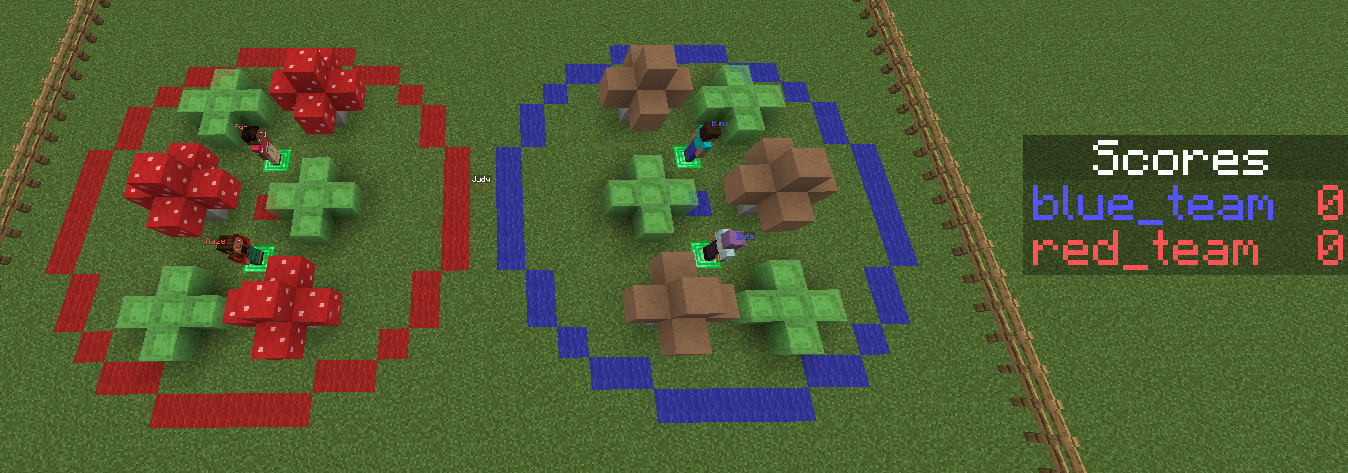}
    \caption{Mushroom War scenario with agents in their starting positions.}
    \label{fig:mushroom-wars}
\end{figure}

In Mushroom War, each team has three small mushrooms, each made up of four mushroom blocks, and three slime patches, each containing four slime blocks (see Figure~\ref{fig:mushroom-wars}). Slime blocks automatically reappear after a random delay when removed, while mushroom blocks only regrow after a random delay if there are seven or fewer slime blocks remaining in the team area. Therefore, it is crucial to continuously remove slime blocks to ensure that mushroom blocks keep regrowing. Harvesting a mushroom block yields 0 to 2 mushrooms, with each mushroom worth 1 point. Only mushrooms collected from a team's own area contribute to their score. The game lasts for 2 minutes, and due to the random delay between removing slime blocks and the regrowth of mushroom blocks, harvesting mushrooms immediately after clearing slime blocks is not advisable. The goal is to outscore the opposing team by the end of the game.

\textbf{Presented Challenge.}
The unique challenge presented in this scenario is task allocation. The key to success is to execute as many actions as possible within the limited time frame. Tasks have to be split between agents in a way that does not get in the way of each other. For example, agents should avoid situations where they are trying to both break the same block. If instead they both break different blocks, they can break twice as many blocks in the same time frame. Finding small optimizations like that is hard and requires a deep understanding of the game mechanics.

\textbf{Built-in opponents} in this scenario encompasses all combinations of 2 sabotage actions that can be done in the game: 1) \textbf{Destroy: }destroying the mushroom blocks in the opponent team area. This prevents the opponent from harvesting the mushroom blocks themselves, 2) \textbf{Place:}  placing slime blocks in the opponent team area. 
Table~\ref{tab:mushroom_war_baselines} shows the strategies of each Mushroom War built-in opponent. Next to sabotaging, the built-in opponents all farm points for their own team. One of the agents is dedicated to removing slime blocks and the other is dedicated to harvesting mushroom blocks. Whenever there are no mushroom blocks to harvest, the mushroom harvesting agent also helps removing slime blocks.
\begin{table}[h!]
\caption{Strategies of Mushroom War built-in opponents with sabotage actions. All built-in opponents remove slime blocks and harvest mushrooms in their own team area.}
\centering
\begin{tabular}{lcc}
\toprule
\textbf{}         & \textbf{Destroy} & \textbf{Place} \\ 
\midrule
\textit{aggressive}  & $\checkmark$            & $\checkmark$            \\ 
\textit{balanced}    & $\checkmark$             &               \\ 
\textit{passive}     &               &               \\ 
\textit{slimy}       &               & $\checkmark$             \\ 
\bottomrule
\end{tabular}
\label{tab:mushroom_war_baselines}
\end{table}

\subsection{Dash \& Dine}
\begin{figure}
    \centering
    \includegraphics[width=0.75\linewidth]{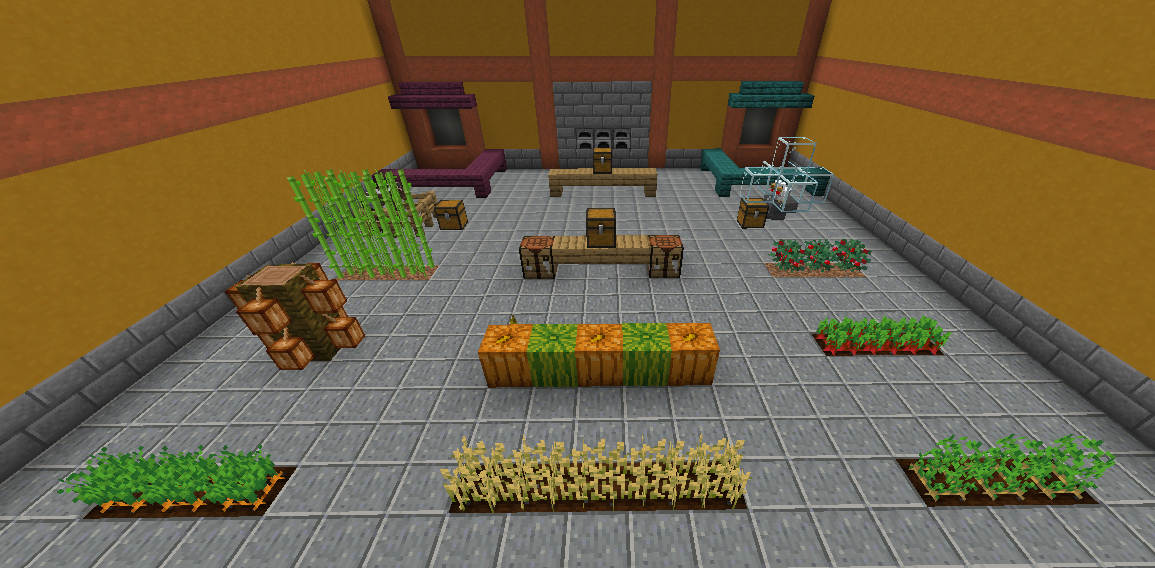}
    \caption{Layout of the Dash \& Dine scenario.}
    \label{fig:dine-and-dash}
\end{figure}
\begin{figure}[t]
    \centering
    \includegraphics[width=1\linewidth]{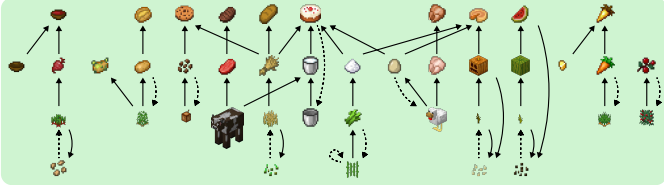}
    \caption{Causal graph of all craftable food items in the Dash \& Dine scenario. There are dependencies between items, blocks, and mobs. Dotted lines represent instancing of new variables, so no variable is the cause of itself but instead a new instance of the same type.}
    \label{fig:dine_full_causal}
\end{figure}
Dash \& Dine is a more complex scenario where agents operate various farms to get ingredients, craft food items, and hand them to their servers to gain points (Figure~\ref{fig:dine-and-dash}). Each team can only submit 3 unique food item types, with any additional food types not awarding any points. Submitted food items give points based on the amount of saturation that the item gives, so more complex and expensive items like pumpkin pie give more points. The arena comes with 8 small farms with different crops, chickens, a cow, crafting tables, 3 pre-fueled furnaces, bowls, buckets, gold nuggets, and a hoe. The random tick rate is set to 200, so crops regrow very quickly, allowing agents to harvest the same farm multiple times. 
The game ends after 2 minutes.
The goal of each team is to win the game. That means having more points than the opposing team at the time the game ends.

\textbf{Presented Challenge.}
The challenge in this scenario lies in the need to plan ahead and adapt to opponent strategies while managing complex subtask dependencies. Players must choose from a variety of food items whose availability depends on the opponent team, and the limit of submitting only three items forces early decisions on which to lock in, all while some food items require multiple crafting steps. Additionally, Dash \& Dine features numerous causal dependencies between items, blocks, and mobs (see Figure~\ref{fig:dine_full_causal}), alongside important spatial and temporal factors. For example, crops require seeds to be planted before they grow to maturity over a random period, with each farm’s location affecting travel time, and smelting takes 10 seconds per item, making it essential to consider processing time when selecting the most point-efficient recipes.
\begin{table}[ht]
\small
\caption{Control Primitives in Mushroom War and Dash \& Dine}
\setlength{\tabcolsep}{1.5pt} 
\centering
\resizebox{0.95\linewidth}{!}{
\begin{tabular}{lp{0.6\linewidth}cc}
\toprule
\makecell[c]{\textbf{Control}\\\textbf{Primitive}} & \textbf{Description} & \makecell[c]{\textbf{Mushroom}\\\textbf{War}} & \makecell[c]{\textbf{Dash \&}\\\textbf{Dine}} \\
\midrule
\textsc{mineBlock}    & Break blocks and pick up the resulting items.  & $\checkmark$ & $\checkmark$ \\
\textsc{craftItem}    & Craft new items from raw materials.           &             & $\checkmark$ \\
\textsc{placeItem}    & Place blocks in the world.                      & $\checkmark$ & $\checkmark$ \\
\textsc{multiAgent}   & Send and receive signals among agents.         & $\checkmark$ & $\checkmark$ \\
\textsc{farm}         & Plant, harvest, or destroy crops.              &             & $\checkmark$ \\
\textsc{smeltItem}    & Smelt items using a furnace.                   &             & $\checkmark$ \\
\textsc{killMob}      & Slay other mobs or players.                     & $\checkmark$ & $\checkmark$ \\
\textsc{giveToPlayer} & Drop items to another player.                   & $\checkmark$ & $\checkmark$ \\
\textsc{useChest}     & Retrieve or deposit items in chests.            &             & $\checkmark$ \\
\textsc{mineflayer}   & Execute basic Mineflayer functions.             & $\checkmark$ & $\checkmark$ \\
\bottomrule
\end{tabular}
}
\label{table:control_primitives}
\end{table}

\textbf{Built-in Opponents.}
The built-in opponents of the Dash \& Dine scenario each target 1 or 2 recipes to craft and hand in to get points, as shown in Table~\ref{tab:opponent_causal_info}.

\begin{table*}[ht]
    \centering
    \small
    \caption{Opponent descriptions and sabotages.}
    \begin{tabular}{l|p{8cm}p{6cm}}
        \hline
        Opponent & Description & Sabotages \\
        \hline
        \textit{berries} & Both agents continuously harvest sweet berries. & Potatoes, beetroots $\rightarrow$ sweet berry bushes. \\
        \hline
        \textit{cake\_beetroot} & One agent crafts cakes from wheat, eggs, milk, and sugar cane; the other crafts beetroot soups. & Melon, pumpkin stems $\rightarrow$ beetroots. \\
        \hline
        \textit{melon\_pumpkin} & One agent continuously harvests melons; the other crafts pumpkin pies from pumpkins, eggs, and sugar cane. & None. \\
        \hline
        \textit{potato\_cookie} & One agent harvests and bakes potatoes; the other crafts cookies from wheat and cocoa beans. & Sweet berry bushes $\rightarrow$ potatoes (using hoe); carrots $\rightarrow$ wheat. \\
        \hline
    \end{tabular}
    \label{tab:opponent_causal_info}
\end{table*}

\section{Methodology - TactiCrafter}

\begin{figure*}
    \centering
    \includegraphics[width=0.8\linewidth]{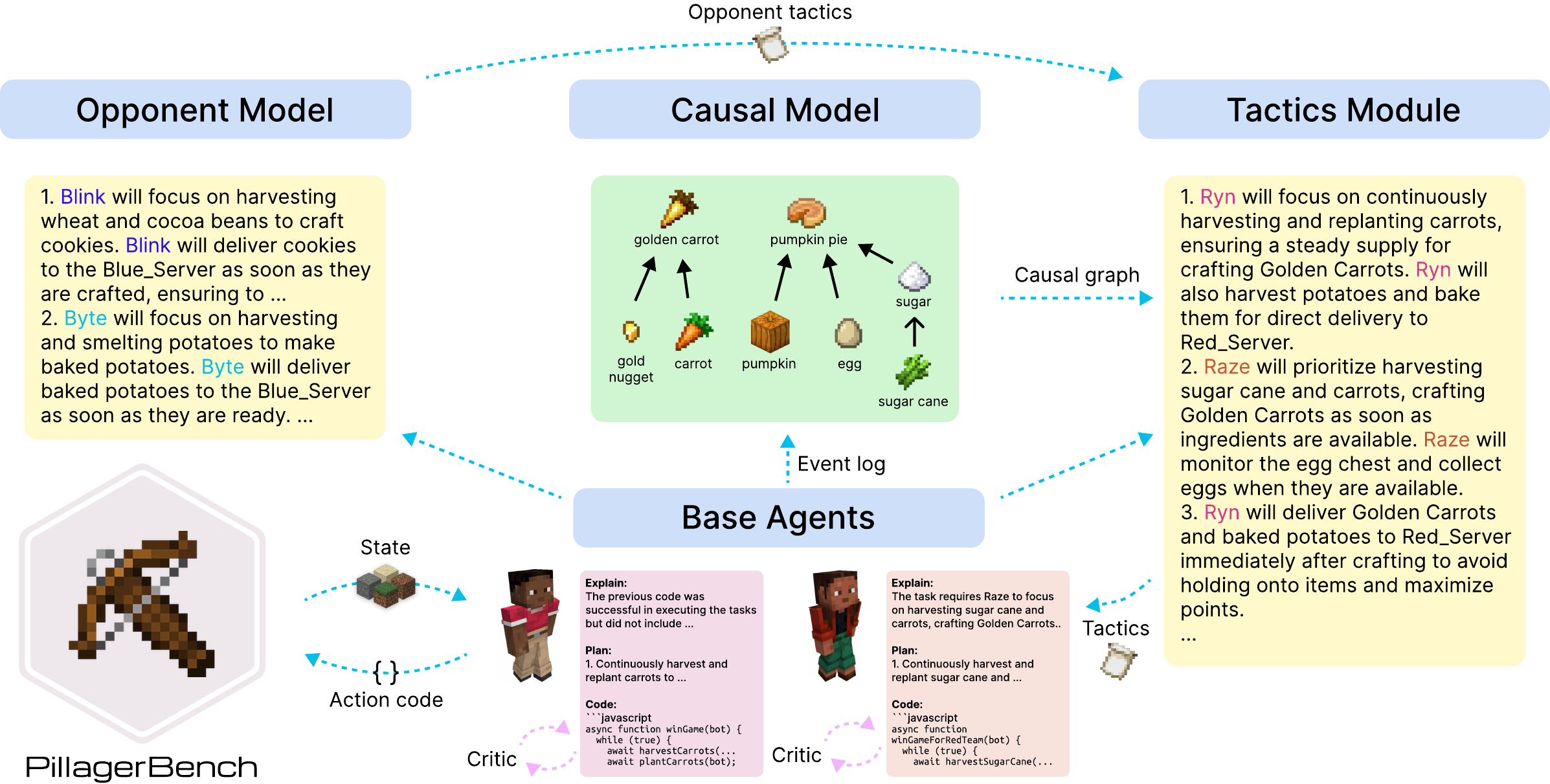}
    \caption{Four key components of TactiCrafter. The \textbf{Causal Model} identifies the causal relationship between items and actions to construct a causal graph. The \textbf{Tactics Module} generates a game-plan that works towards the objective and counters the opponent tactics. The \textbf{Opponent Model} identifies actions of opponent agents and tries to guess their tactics. The \textbf{Base Agents} execute actions in the environment according to the tactics and records the events.}
    \label{fig:tacticrafter}
\end{figure*}

In this section, we introduces a LLM-based multi-agent system called TactiCrafter, 
which comprises four main components: Tactics Module, Causal Model, Opponent Model, and Base Agents.
It operates by having the Tactics Module generate tactics each episode, based on the scenario description, causal graph, opponent tactics, and events from last round. Then the Base Agents will carry out the execution of these tactics and self-reflection.
Afterwards the recorded events of the episode will be used to update the Causal Model and Opponent Model, which are responsible for maintaining information about how the world works and the behavior of the opponent teams.
See Figure~\ref{fig:tacticrafter} for a general outline.

\subsection{Tactics Module}
The Tactics Module is responsible for generating the tactics for the team, which is a high level description of the action policy in natural language. A human can thus read this to very quickly figure out what TactiCrafter is trying to accomplish (Figure~\ref{fig:tacticrafter}). Inspired by the contracts in ~\cite{yocum2023mitigating}, our tactics is a list conditional actions between players, consisting of numbered short, simple, specific Tactics line items (no more than 6 lines), where players are referenced by name in third person.

\paragraph{Initial Tactics Generation}
In the first episode, the Tactics Module is tasked to generate the initial tactics $\mathbb{T}$. Given the game description $D$ (team name, team objective, scenario description), and causal graph $\mathbb{G}$, the Tactics Module generates the tactics $\mathbb{T}$:
\begin{align}
    \mathbb{T} &= \text{LLM}(p_a,D,\mathbb{G})
\end{align}
where $p_a$ is the initial tactics prompt which adopts a zero-shot chain-of-thought (CoT) approach~\cite{wei2023chainofthoughtpromptingelicitsreasoning}.
The tactics have to be generated between special tags which indicate the begin and end of the tactics.

\paragraph{Tactics Update}
The Tactics Module is also used to update the tactics in subsequent episodes. Given the game description $D$ (team name, team objective, scenario description), historical information $H$ (previous events, previous tactics), causal graph $\mathbb{G}$, and opponent tactics $\mathbb{O}$, the Tactics Module generates the updated tactics $\mathbb{T}$:
\begin{align}
    \mathbb{T} &= \text{LLM}(p_b,D,H,\mathbb{G},\mathbb{O})
\end{align}
where $p_b$ is the tactics update prompt which adopts a few-shot CoT approach~\cite{wei2023chainofthoughtpromptingelicitsreasoning}.
From the previous events we extract the chat log, nearby blocks, and inventory of all team members.

\subsection{Causal Model}
The scenario description describes most of the necessary information about how the game and the world work, like required crafting recipes, but it can leave out details by being concise, leaving opportunities to learn hidden mechanics.
The Causal Model is responsible for compiling information about how the world works by creating a causal graph using LLM-based causal discovery.
The causal graph defines for each action which items are necessary to cause that action and which items are a result of that action.
Inspired by the success of Adam~\cite{yu2024adamembodiedcausalagent} we added the Causal Model from their paper and modified it to be able to work within PillagerBench. To work within PillagerBench's limited execution time, we opted to remove the intervention module which meticulously verifies every causal relation by way of interventions. Instead we have to rely on identifying causal correlations from observations.

The causal graph $\mathbb{G}$ is formatted into a text representation by encoding each causal relation as a 3-tuple: an action, a list of its causes, and a list of its effects. For example: \texttt{Action: craftItem(bot, "bread", 5); Cause: ['wheat']; Effect ['bread']}\\
Since actions in PillagerBench are code, actions in the causal graph are represented as code snippets which invoke a single function with specific arguments.

\paragraph{Initial Causal Graph Generation}
In the first episode, the Causal Model is tasked to generate the initial causal graph $\mathbb{G}$. Given the game description $D$ (control primitives context, scenario description), the Causal Model generates the initial causal graph $\mathbb{G}$:
\begin{align}
    \mathbb{G} &= \text{LLM}(p_c,D)
\end{align}
where $p_c$ is the initial causal discovery prompt which utilizes one-shot prompting. The LLM outputs in JSON at least one causal relationship for every control primitive. The causal discovery is based on the LLMs understanding of the code in the control primitives context and scenario description.

\paragraph{Causal Graph Update}
In subsequent episodes, the Causal Model uses observations from the last episode to update the causal graph. Given the game description $D$ (control primitives context, scenario description), historical information $H$ (previous events), and causal graph $\mathbb{G}$, the Causal Model generates a set of new causal relations $\mathbb{G'}$:
\begin{align}
    \mathbb{G'} &= \text{LLM}(p_d,D,H,\mathbb{G})\\
    \mathbb{G} &= \mathbb{G'} \cup \mathbb{G}
\end{align}
where $p_d$ is the causal discovery update prompt which utilizes few-shot prompting. 
From the previous events we extract a list of tuples which describes for every team member and for every chat message they sent, the content of the message and inventory state at the time.
From the chat messages, the executed actions must be inferred, and from the inventory states the causes and effects must be inferred.
Sometimes the chat messages contain additional environment feedback which can be helpful for inferring cause and effect. For example a failed invocation of \textsc{craftItem} can say: ``I cannot make bread because I need: 2 more wheat'' which immediately tells you that the action is crafting bread and it requires wheat.
Agents are encouraged to broadcast all their actions via chat messages. Additionally, their usage of control primitives is automatically broadcast. All these messages are visible to opponents.

\subsection{Opponent Model}
The Opponent Model is responsible for figuring out what the opponent team is doing and summarizing it as tactics. After every episode, the Opponent Model uses the chat log of the opponent team to update its tactics hypothesis. Given the the game description $D$ (opponent team name, opponent team objective, scenario description), historical information $H$ (previous events, previous opponent tactics), and causal graph $\mathbb{G}$, the Opponent Model generates the updated opponent tactics $\mathbb{O}$:
\begin{align}
    \mathbb{O} &= \text{LLM}(p_e,D,H,\mathbb{G})
\end{align}
where $p_e$ is the opponent tactics update prompt which adopts a few-shot CoT approach~\cite{wei2023chainofthoughtpromptingelicitsreasoning}.
From the previous events we extract the chat messages coming from players in the opposing team only.
In the first iteration, $\mathbb{O'}$ is initialized as ``unknown''.

\subsection{Base Agents}

The Base Agents are responsible for carrying out the tactics and actually interacting with the PillagerBench environment. There is one Base Agent per player in the team and they work in parallel.
It uses the iterative prompting mechanism from Voyager~\cite{wang2023voyageropenendedembodiedagent} to self-improve by incorporating feedback and execution errors from the PillagerBench environment and self-critique about about tactics adherence.

The Base Agent iterates a sequence of generating action code, executing the code, and self-criticizing.
Given the game description $D$ (agent username, control primitives context, scenario description, team objective), historical information $H$ (tactics from previous episode, code from previous iteration, execution error from previous iteration, events from previous iteration), critique $\mathbb{C}$, and tactics $\mathbb{T}$, the action agent generates new action code $\mathbb{A}$:
\begin{align}
    \mathbb{A} &= \text{LLM}(p_f,D,H,\mathbb{C},\mathbb{T})
\end{align}
where $p_f$ is the action template prompt which adopts a zero-shot CoT approach~\cite{wei2023chainofthoughtpromptingelicitsreasoning}.

Given the game description $D$ (agent username, team objective, scenario description), historical information $H$ (execution error from the previous iteration, events from the previous iteration), and tactics $\mathbb{T}$, the critic agent generates critique for the action agent $\mathbb{C}$:
\begin{align}
    \mathbb{C} &= \text{LLM}(p_g,D,H,\mathbb{T})
\end{align}
where $p_g$ is the critic prompt which adopts a few-shot CoT approach~\cite{wei2023chainofthoughtpromptingelicitsreasoning}. From the previous events we extract the chat log, biome, time, nearby blocks, nearby entities, health, hunger, position, and inventory for both the action and critic agents.

\paragraph{Event Logging}
Each Base Agent records their own log of events from the state information they obtain from PillagerBench by concatenating the state information after every action roll-out iteration. At the end of the episode we select the Base Agent with the longest (most complete) event log to give to the Tactics Module and Opponent Model. The Causal Model simply concatenates the chat events sent by each Base Agent in the team.

\paragraph{Event Deduplication}
With code as actions in PillagerBench, it is quite common for the code to contain an infinite loop that repeatedly attempts a sequence of actions. This results in event logs that contain the same sequence of chat messages many times while not bringing any extra valuable information. To save costs on input tokens, we therefore deduplicate these event logs with an algorithm that removes any repeated sequences of chat messages. This reduces the total number of events by 28\% when running TactiCrafter on PillagerBench. 

\section{Experiments}
In this section we give the implementation details and methodology for testing and evaluating agents, including performance metrics, results, and discussion.

\subsection{Setup}
\paragraph{Implementation Details}
The PillagerBench benchmark is configured with the Mushroom War and Dash \& Dine scenarios. Both scenarios use 2 teams (red team \& blue team) with 2 agents each, and each episode lasts 2 minutes. Environment wait ticks is set to 80.
The Mushroom War scenario has the \textit{do\_nothing}, \textit{aggressive}, \textit{balanced}, \textit{passive}, and \textit{slimy} built-in opponents.
The Dash \& Dine scenario has the \textit{do\_nothing}, \textit{berries}, \textit{cake\_beetroot}, \textit{melon\_pumpkin}, and \textit{potato\_cookie} built-in opponents.
For reporting results we divide the points in the Dash \& Dine scenario by 10 to bring it in line with the average point gain in Mushroom War.

A typical full benchmark run has the multi-agent system of choice play against each built-in opponent of every scenario for 5 consecutive episodes. We repeat this 3 times.

We evaluate the performance of TactiCrafter, a novel LLM-based competitive multi-agent system, on PillagerBench. 
LLM responses for TactiCrafter with the GPT-4o\footnote{\texttt{gpt-4o-2024-08-06}} model~\cite{gpt4o} are provided by the OpenAI Chat Completions API endpoint, and responses for Gemini 2.0 Flash\footnote{\href{https://openrouter.ai/google/gemini-2.0-flash-001}{https://openrouter.ai/google/gemini-2.0-flash-001}} and o3 Mini\footnote{\href{https://openrouter.ai/openai/o3-mini}{https://openrouter.ai/openai/o3-mini} (medium reasoning effort)} are provided by OpenRouter~\cite{openrouter}. We use sampling temperature 0.3 in all experiments and modules. When comparing different LLM models for TactiCrafter we replace the LLM model in all modules and run the full benchmark.
In all other experiments we use the GPT-4o\footnotemark[1] LLM for inferences.

\paragraph{Baselines}
To prove the effectiveness of TactiCrafter we will compare it against these baselines on PillagerBench:
\begin{enumerate}
    \item \textbf{Random Baseline:} This baseline has all agents in the team continuously perform random actions. It generates executable action code by randomly selecting a control primitive or Mineflayer function and inserting random arguments that choose from nearby blocks, nearby entities, inventory items, or players. Simple checks are performed to make sure the actions are executable before executing them.
    \item \textbf{CoT Baseline:} This is a simple LLM-based multi-agent system that uses a single Chain-of-Thought~\cite{wei2023chainofthoughtpromptingelicitsreasoning} prompt to generate executable action code for all agents in the team. It does this once per episode in the pre-game step. It utilizes the same GPT-4o LLM as TactiCrafter.
\end{enumerate}
Given the game description $D$ (nearby blocks, nearby entities, team name, team agents, scenario description, team objective), and historic information $H$ (code from last episode, execution error from last episode, chat log from last episode), the CoT Baseline generates new action code $\mathbb{A}_1, \mathbb{A}_2,\dots$:
\begin{align}
    \mathbb{A}_1, \mathbb{A}_2,\dots &= \text{LLM}(p_h,D,H)
\end{align}
where $p_h$ is the CoT Baseline action prompt which adopts a zero-shot CoT approach~\cite{wei2023chainofthoughtpromptingelicitsreasoning}.

\paragraph{Evaluation Metrics}
We evaluate the multi-agent system using 4 key metrics: 1) \textbf{Points ($P$)} measures the average points scored in game. 2) \textbf{Sabotage ($S$)} measures the average points denied from the opponent team. 3) \textbf{Points difference ($D$)} measures the average point difference w.r.t. opponent teams. 4) \textbf{Win rate ($W$)} measures the ratio of game episodes won.

\small
\begin{equation}
    \begin{array}{lll}
    P &= \frac{1}{N_e} \sum_{i=1}^{N_e} S_{\text{red}, i}, \quad
    S = \sigma_{\text{blue}} - \frac{1}{N_e} \sum_{i=1}^{N_e} S_{\text{blue}, i}, \\
    D &= \frac{1}{N_e} \sum_{i=1}^{N_e} (S_{\text{red}, i} - S_{\text{blue}, i}), \\
    W &= \frac{1}{N_e} \sum_{i=1}^{N_e} ( 
        \mathbf{1}(S_{\text{red}, i} > S_{\text{blue}, i}) + 0.5 \cdot \mathbf{1}(S_{\text{red}, i} = S_{\text{blue}, i}))
    \end{array}
\end{equation}

\normalsize
Here $N_e$ is the number of episodes, $S_{\text{red}, i}$ is the score achieved by the red team at the end of episode $i$, and $S_{\text{blue}, i}$ is the score achieved by the blue team at the end of episode $i$.
$\sigma_{\text{blue}}$ is the average score achieved by the blue team if the red team is the \textit{do\_nothing} built-in opponent. $\sigma_{\text{blue}}$ is calculated in advance by running simulations of every built-in opponent match-up for 20 episodes.

For fair comparison between LLMs we will also measure the average \textbf{response time ($T_{resp}$)}, \textbf{number of output tokens ($N_{out}$)}, \textbf{tokens per second ($R_{tps}$)}, and \textbf{Roll-out iterations per episode ($I$)}. The response time can influence the performance a lot since PillagerBench is a real-time benchmark, and the response time can vary a lot between providers. Also reasoning models have longer response times because they generate way more tokens when including reasoning tokens. If there is an execution error and the code has to be re-evaluated, the base agent is not doing anything while the LLM is generating new code, so it is advantageous to generate responses quicker.

Roll-out iterations counts how many times code has been generated and executed within the same episode. This is always at least one, but the first roll-out iteration does not contribute to additional waiting time for the agent during the game phase, because the code has been generated during the pre-game phase. Therefore the expected idle time during the game phase is: $T_{\text{resp}}\cdot (I-1)$

\small
\begin{equation}
    \begin{array}{ll}
T_{\text{resp}} &= \frac{1}{N_{\text{llm}}} \sum_{i=1}^{N_{\text{llm}}} T_{\text{resp}, i}, \qquad 
N_{\text{out}}  = \frac{1}{N_{\text{llm}}} \sum_{i=1}^{N_{\text{llm}}} N_{\text{out}, i}, \\[1ex]
R_{\text{tps}}  &= \frac{1}{N_{\text{llm}}} \sum_{i=1}^{N_{\text{llm}}} \frac{N_{\text{out}, i}}{T_{\text{resp}, i}}, \qquad 
I             = \frac{1}{N_e} \sum_{i=1}^{N_e} I_{i}.
    \end{array}
\end{equation}
\normalsize
Here $N_{\text{llm}}$ is the number of LLM responses, $T_{\text{resp}, i}$ is the time in seconds required to generate LLM response $i$, $N_{\text{out}, i}$ is the number of generated tokens in LLM response $i$, and $I_{i}$ is the number of roll-out iterations in episode $i$.

\subsection{Benchmark Analysis}
We compare the built-in opponents against each other and evaluate 40 episodes for Mushroom War, 20 episodes for Dash \& Dine.

\begin{figure*}
    \centering
    \includegraphics[width=1\linewidth]{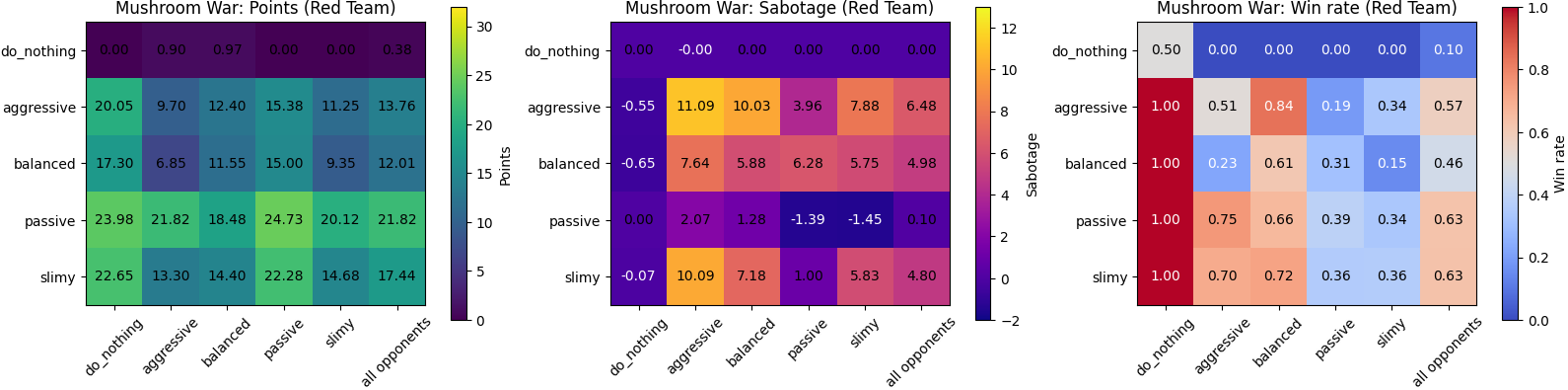}
    \caption{Average points ($P$), sabotage ($S$), and win rate ($W$) for each built-in opponent matchup in the Mushroom War scenario. The labels on the left side denote the agent playing the red team, and the labels on the bottom side denote the agent playing the blue team.}
    \label{fig:mushroom_war_baseline}
\end{figure*}

\begin{figure}
    \centering
    \includegraphics[width=1\linewidth]{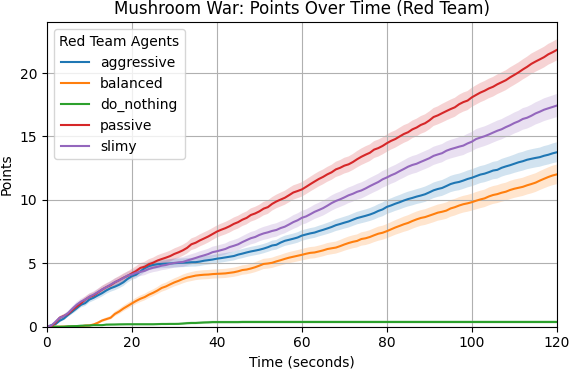}
    \caption{Average points ($P$) against all opponents for each built-in opponent plotted over time of a 2-minute episode of the Mushroom War scenario. The shaded area shows the 5\% confidence interval.}
    \label{fig:mushroom_war_baseline_timeline}
\end{figure}

\textbf{Built-in opponents have highly varied strategies.} In Figure~\ref{fig:mushroom_war_baseline}, the \textit{passive} opponent scores highest in Points, with \textit{slimy} close in win rate due to its sabotage ability despite fewer points. Opponents that break enemy mushrooms score high in sabotage but low in team points—likely because sabotage takes time and dropped mushrooms may be collected by opponents. This also explains the non-zero points and negative sabotage score for the \textit{do\_nothing} agent in Figure~\ref{fig:mushroom_war_baseline_timeline}.

\begin{figure*}
    \centering
    \includegraphics[width=1\linewidth]{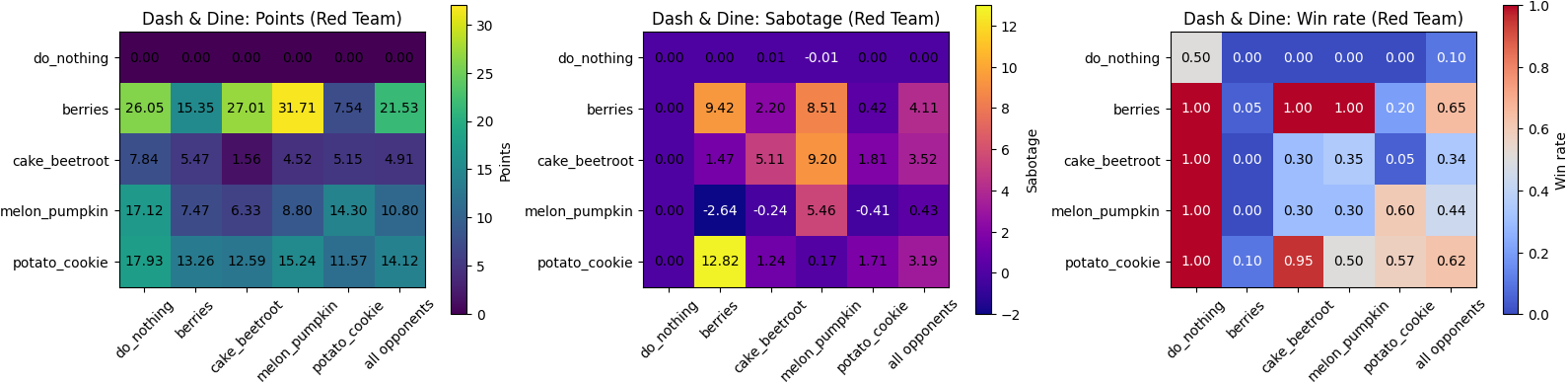}
    \caption{Average points ($P$), sabotage ($S$), and win rate ($W$) for each built-in opponent matchup in the Dash \& Dine scenario. The labels on the left side denote the agent playing the red team, and the labels on the bottom side denote the agent playing the blue team.}
    \label{fig:dash_and_dine_baseline}
\end{figure*}

\begin{figure}
    \centering
    \includegraphics[width=1\linewidth]{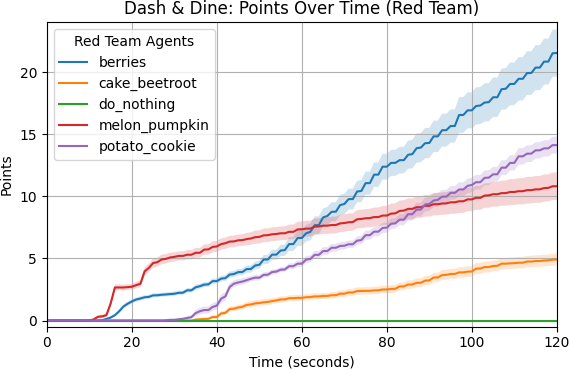}
    \caption{Average points ($P$) against all opponents for each built-in opponent plotted over time of a 2-minute episode of the Dash \& Dine scenario. The shaded area shows the 5\% confidence interval.}
    \label{fig:dash_and_dine_timeline}
\end{figure}

Figure~\ref{fig:dash_and_dine_baseline} shows the \textit{berries} opponent leading in points, sabotage, and win rate—losing only to \textit{potato\_cookie}, which targets sweet berry bushes (as the sabotage heatmap indicates). Its advantage stems from sweet berries' rapid regrowth and decent saturation, along with its unique ability to convert multiple crop types into sweet berry bushes, which boosts points over time (see Figure~\ref{fig:dash_and_dine_timeline}). Additionally, the arena's asymmetry places sweet berry bushes closer to the blue team, giving them an edge in self-play.

It is also noteworthy in Figure~\ref{fig:dash_and_dine_timeline} that the time until the first points scored are a lot longer than in Figure~\ref{fig:mushroom_war_baseline_timeline}. This shows how Dash \& Dine is a more complicated scenario than Mushroom War, requiring more consecutive steps to gain points. Also the built-in opponents that use more complex recipes like cake, beetroot soup, or baked potatoes show even more delay in gaining their first points.

\subsection{Main Results}

\begin{figure*}
    \centering
    \includegraphics[width=1\linewidth]{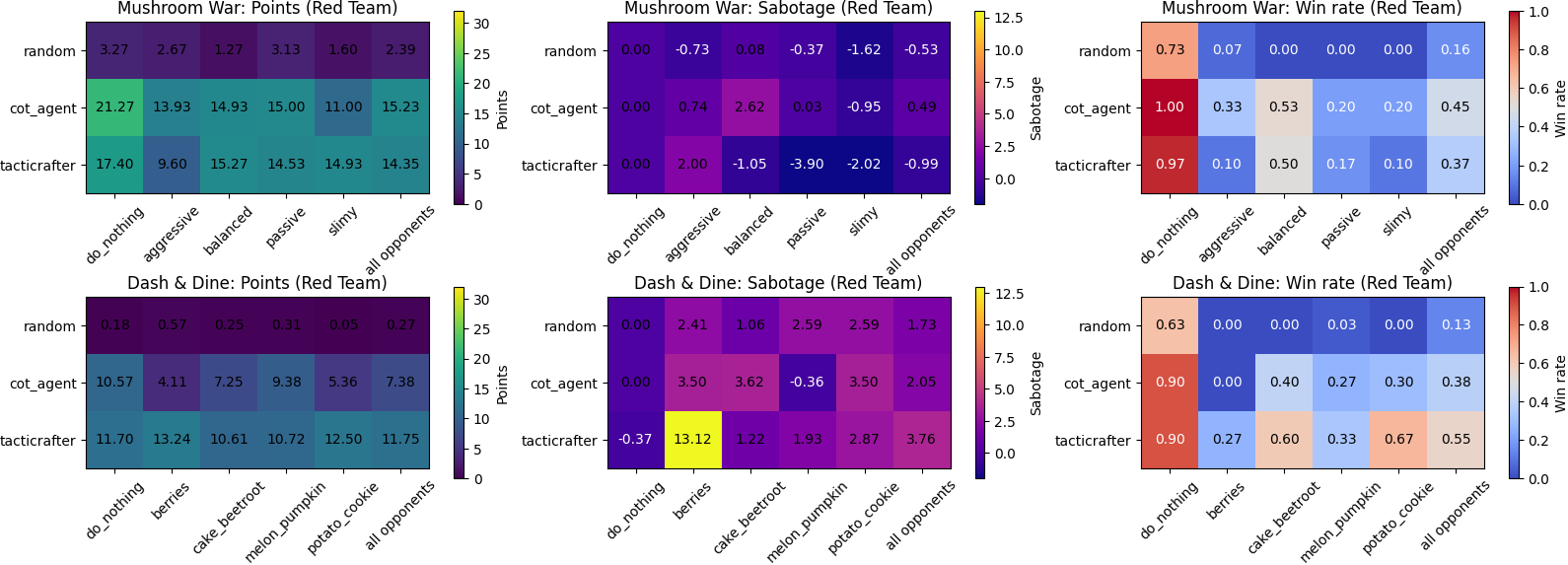}
    \caption{Average points ($P$), sabotage ($S$), and win rate ($W$) for each baseline and TactiCrafter against all built-in opponents. The top row shows Mushroom War and the bottom row shows Dash \& Dine. The labels on the left side denote the agent playing the red team, and the labels on the bottom side denote the agent playing the blue team.}
    \label{fig:baselines_details}
\end{figure*}

In Table~\ref{tab:baselines} we have laid out the performance metrics of all baselines and our proposed method TactiCrafter. TactiCrafter achieves the highest points, sabotage, points difference, and win rate, showing superiority over the baselines. The Chain-of-Thought (CoT) Baseline is not far behind TactiCrafter and it performs well in the Mushroom War scenario, but not as well in the Dash \& Dine scenario (Figure~\ref{fig:baselines_details}). This means that the main advantage of TactiCrafter comes from solving complex dynamic scenarios where opponent strategy plays a big role.
The Random Baseline performs terribly with a near zero win-rate against all but the \textit{do\_nothing} built-in opponents. This is expected because the action space in PillagerBench is very large, so the probability of picking productive actions is low.

Despite the awesome achievements of TactiCrafter, the win rate against built-in opponents is still low, so there is lots of room for improvement. The built-in opponents have the disadvantage of not being able to adapt to their opponents, so theoretically it should be possible to get a greater than or equal win rate with a more advanced multi-agent system that \textit{is} able to adapt to their opponents.

\begin{table}[t]
    \caption{Comparison of TactiCrafter versus the Random Baseline and the Chain-of-Thought (CoT) Baseline on PillagerBench. TactiCrafter achieves the highest points, sabotage, points difference, and win rate.}
    \centering
    \begin{tabular}{l l l l l}
        \toprule
        \textbf{Baseline} & $P\uparrow$ & $S\uparrow$ & $D\uparrow$ & $W\uparrow$\\
        \midrule
        Random & 1.33 & 0.67 & -13.66 & 0.15\\
        CoT & 11.33 & 1.41 & -2.88 & 0.42\\
        TactiCrafter & \textbf{13.05} & \textbf{1.55} & \textbf{-1.16} & \textbf{0.46}\\
        \bottomrule
    \end{tabular}
    \label{tab:baselines}
\end{table}

\subsection{Ablation Study}
In this section we test several ablations of TactiCrafter to discover the optimal configuration of hyperparameters.

\textbf{Ablation on LLM backbones.}
We compare GPT-4o, Gemini 2.0 Flash, and o3 Mini medium for TactiCrafter by replacing the LLM model in all modules.
In Table~\ref{tab:model_comparison} we can see GPT-4o achieves the best performance, followed by Gemini 2.0 Flash.  Gemini 2.0 Flash also has the highest number of action iterations per episode which is offset by having the highest tokens-per-second rate and fastest response time. Lastly o3 Mini achieves the lowest winrate, but it manages to most effectively sabotage the opponent team. This means it focuses more on sabotaging the opponent team rather than obtaining points for their own team, at the detriment to its own score.

\begin{table}[t]
\small
\setlength{\tabcolsep}{2pt} 
    \caption{Abalation on different LLM backbones.}
    \centering
    \begin{tabular}{l l l l l l l l l}
        \toprule
        \textbf{Model} & $P\uparrow$ & $S\uparrow$ & $D\uparrow$ & $W\uparrow$ & $T_{\text{resp}}\downarrow$ & $N_{\text{out}}$ & $R_{\text{tps}}\uparrow$ & $I\downarrow$\\
        \midrule
        gpt-4o   & \textbf{13.05} & 1.55 & \textbf{-1.16} & \textbf{0.46} & 9.16 & 390.39 & 42.64 & \textbf{1.35}\\
        gemini   & 12.10        & 1.00 & -2.60       & 0.40         & \textbf{5.36} & 731.15 & \textbf{136.40} & 1.51\\
        o3-mini  & 7.91         & \textbf{2.44} & -5.50   & 0.36         & 25.45 & 2805.31 & 110.24 & 1.37\\
        \bottomrule
    \end{tabular}
    \label{tab:model_comparison}
\end{table}

\begin{figure}
    \centering
    \includegraphics[width=1\linewidth]{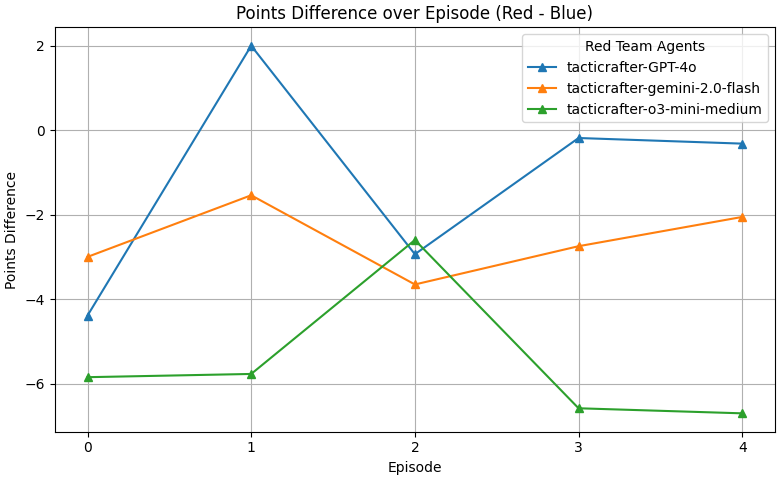}
    \caption{Evolution of average reward difference ($D$) over 5 consecutive episodes for TactiCrafter on PillagerBench with different LLMs.}
    \label{fig:D_episode}
\end{figure}

We have plotted the average reward difference ($D$) over 5 consecutive episodes for TactiCrafter on PillagerBench with different LLMs in Figure~\ref{fig:D_episode}. Each model displays a peak in performance around episode 1 or 2 followed by a big dip in performance. GPT-4o and Gemini 2.0 Flash both achieve higher rewards in the last episode compared to the first episode, while o3-mini ends up with lower performance in the last episode. GPT-4o shows the strongest upwards trend in performance over episode, suggesting a good ability to improve over time as it gains more experiences.

\textbf{Ablation on TactiCrafter Modules.}
To find out which modules really contribute to the success of TactiCrafter, we run an ablation study for the Causal Model and the Opponent Model. For this, we take TactiCrafter using GPT-4o as backing LLM and disable either the Causal Model or the Opponent Model and then do a full run on the PillagerBench benchmark.

\begin{table}[t]
\small
\setlength{\tabcolsep}{2pt} 
    \caption{Ablation on the components of TactiCrafter on PillagerBench. }
    \centering
    \begin{tabular}{l l l l l l}
        \toprule
        \textbf{Ablation} & $P\uparrow$ & $S\uparrow$ & $D\uparrow$ & $W\uparrow$ & $I\downarrow$\\
        \midrule
        TactiCrafter & \textbf{13.05} & 1.55 & -1.16 & \textbf{0.46} & 1.35\\
        TactiCrafter w/o causal & 12.80 & 2.08 & \textbf{-0.92} & 0.44 & \textbf{1.31}\\
        TactiCrafter w/o opponent & 11.83 & \textbf{2.32} & -1.68 & 0.45 & 1.42\\
        \bottomrule
    \end{tabular}
    \label{tab:ablation}
\end{table}

The results of the ablation study are shown in Table~\ref{tab:ablation}. The TactiCrafter achieves the highest win rate by a small margin, TactiCrafter without the causal model achieves the best points difference, and TactiCrafter without the opponent model is the best at sabotaging the opponent team.
The TactiCrafter has the lowest sabotage score and highest points score. Therefore, perhaps the additional knowledge provided by the opponent module and the causal module help shift focus to a more defensive play style that focuses more on maximizing individual point gain and defending against opponent team sabotages.

\textbf{TactiCrafter learns to adapt to specific opponents.}
To test the adaptability of TactiCrafter we let TactiCrafter play against an built-in opponent for 5 consecutive episodes and make a checkpoint of the TactiCrafter experiences. Then we run TactiCrafter from that checkpoint for a 6th episode against every built-in opponent in the scenario, starting from the same checkpoint every time. We then record the performance with the same opponent vs. a different opponent. We repeat this for all built-in opponents and scenarios, and then repeat that 3 times.

We observe a strong adaptation in TactiCrafter to the specific opponent team's strategy after playing one opponent for 5 episodes. The results are shown in Table~\ref{tab:adaptation}. When playing the same opponent again, TactiCrafter obtains more points, denies the opponent more points, and achieves a higher win rate than if it were to play a random different opponent.
The points difference achieved by TactiCrafter when playing a different opponent is much lower than that if it were playing the same opponent, however it is still higher than the point difference achieved on the first episode with no prior experience. This means some of the knowledge gained while playing those 5 episodes against the same opponent generalizes to playing other opponents too.
Interestingly enough, TactiCrafter does not observe a change in win rate in Mushroom War when faced with a different opponent, suggesting that specializing to a specific opponent plays a smaller role in that scenario.

\begin{table}
    \caption{Performance of TactiCrafter on PillagerBench (All scenarios, Mushroom War (MW), Dash \& Dine (DD)) when faced with the same or a random different built-in opponent after having played the same opponent for 5 episodes. }
    \centering
    \begin{tabular}{l l l l l}
        \toprule
        \textbf{Opponent} & $P\uparrow$ & $S\uparrow$ & $D\uparrow$ & $W\uparrow$\\

        \midrule
        MW Same & \textbf{15.53} & \textbf{1.69} & \textbf{0.33} & 0.33\\
        MW Different & 13.95 & -0.71 & -3.65 & 0.33\\
        \midrule
        DD Same & \textbf{11.12} & \textbf{5.91} & \textbf{2.73} & \textbf{0.60}\\
        DD Different & 10.32 & 1.94 & -2.04 & 0.44\\
        \midrule
        Avg Same & \textbf{13.33} & \textbf{3.80} & \textbf{1.53} & \textbf{0.47}\\
        Avg Different & 12.14 & 0.61 & -2.84 & 0.38\\
        \bottomrule
    \end{tabular}
    \label{tab:adaptation}
\end{table}

\textbf{Self-play can improve overall performance if opponent specialization is not a factor.}
We let TactiCrafter play against an identical instance of itself for 20 consecutive episodes in each scenario repeated 3 times. Every 5 episodes we made a checkpoint and evaluated the performance of the red team TactiCrafter by letting it play against all built-in opponents for one episode.

\begin{figure}
    \centering
    \includegraphics[width=1\linewidth]{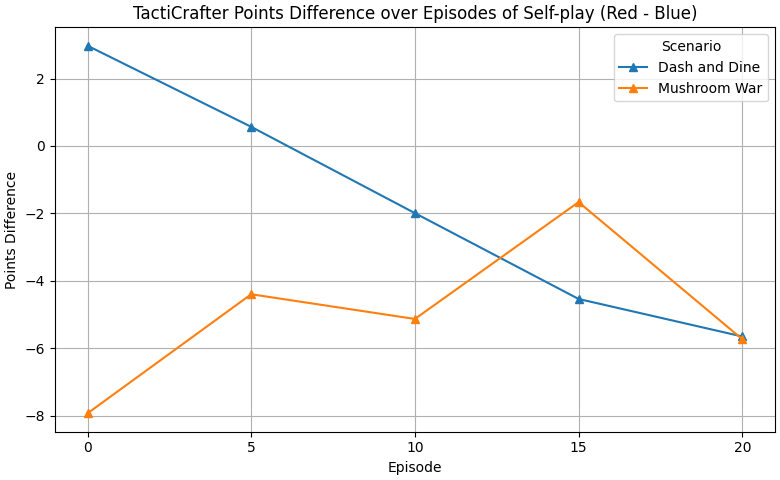}
    \caption{The performance of TactiCrafter in PillagerBench after a number of episodes of self-play.}
    \label{fig:self_play_improvement}
\end{figure}

As shown in Figure~\ref{fig:self_play_improvement}, there is an improvement over time in Mushroom War, going from -8 to -1.8 points difference after 15 episodes of self-play, however in Dash \& Dine it gets repeatedly worse, going from 3 to -5.2 points difference after 20 episodes.
This difference can be explained by opponent strategy being a much more important factor in Dash \& Dine. As TactiCrafter gets more specialized against its TactiCrafter opponent, a sudden switch to a built-in opponent with a different strategy has a huge impact on performance. Whereas in Mushroom War, the opponent strategy has little impact on your own objectives, so improvements in action efficiency from self-play experience can actually cause an improvement in points difference against built-in opponents.

In self-play, TactiCrafter overfits to its own strategies, leading to worse performance when faced with a different opponent. This raises an important question: how can we maintain adaptability while preventing detrimental overspecialization?
To improve this, we could execute a protocol that removes the specialization information from TactiCrafter while keeping the objective knowledge of the world. This means resetting the Opponent Model and keeping the information in the Causal Model. The tactics could also be updated in a fashion that takes into account that the opponent is unknown. For example, removing sabotage actions because we do not know yet which sabotages will be effective in disrupting the opponent.

\subsection{Case Study}

\begin{figure}
    \centering
    \includegraphics[width=1\linewidth]{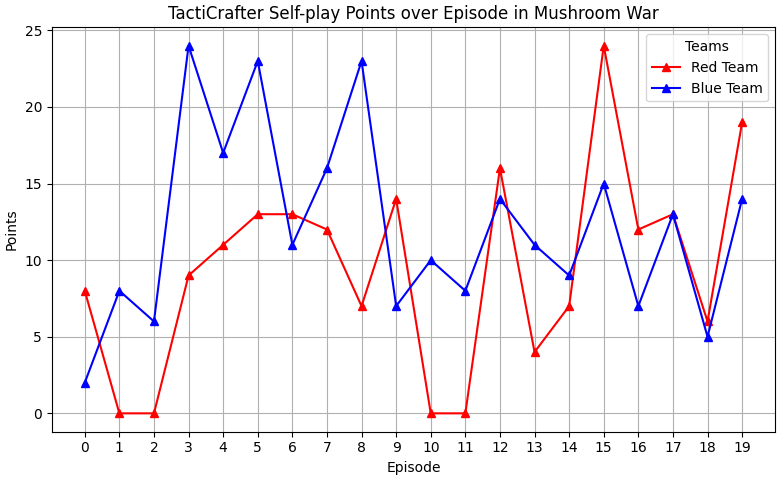}
    \caption{Points of the red team and the blue team for a game of 20 consecutive episodes of Mushroom War. Both teams are played by instances of TactiCrafter.}
    \label{fig:selfplay_mushroom_war}
\end{figure}

\begin{figure}
    \centering
    \includegraphics[width=1\linewidth]{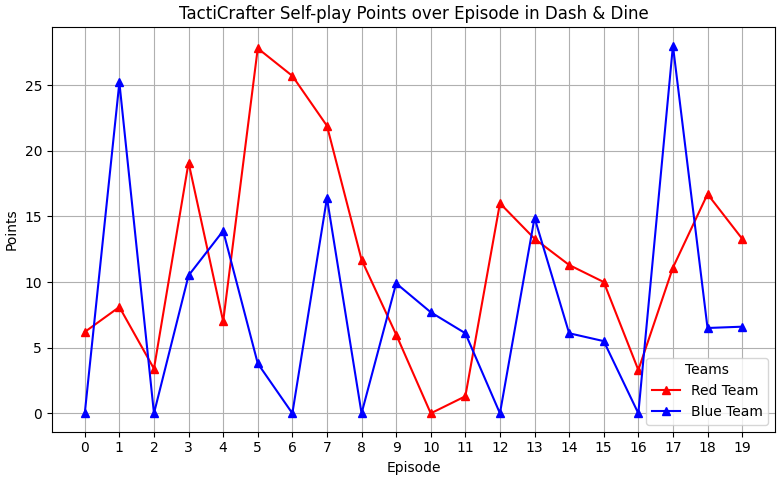}
    \caption{Points of the red team and the blue team for a game of 20 consecutive episodes of Dash \& Dine. Both teams are played by instances of TactiCrafter.}
    \label{fig:selfplay_dash_and_dine}
\end{figure}
\textbf{Self-play incites evolving tactics.} We report the performance of TactiCrafter for Mushroom War in Figure~\ref{fig:selfplay_mushroom_war} and Dash \& Dine in Figure~\ref{fig:selfplay_dash_and_dine}. In first 3 episodes both teams did not manage to score above 10 points, however afterwards they managed to score above 10 points about 65\% of the time. We can also observe the blue team being dominant in the first 12 episodes and afterwards the red team taking the lead, suggesting a shift in strategies that leveled the playing field.

As we can see frequent big shifts in points between episodes, even hitting zero points quite often in the later episodes. There is a brief period around episode 5 where the red team did exceptionally well with a 26 point lead over the blue team. Let us have a closer look at what was going on in this episode:
\begin{itemize}
    \item \textbf{\textcolor{red}{The red team:}} Ryn continuously harvests and replants carrots and harvests sweet berries and delivers these to the server. Raze continuously harvests sweet berries and melons and delivers these to the server.
    \item \textbf{\textcolor{blue}{The blue team:}} Blink incorrectly assumes there are no sweet berries so it moves to harvest raw potatoes without replanting and delivers them to the server. Byte harvests potatoes without replanting, destroys sweet berry bushes, smelts 2 potatoes and delivers these to the server.
\end{itemize}
The blue team had a solid strategy to bake potatoes and harvest sweet berries but failed in the execution. Blink failed to check for sweet berries because the action code checked the inventory for the presence of sweet berries instead of the environment, and both failed to replant potatoes after harvesting them, effectively destroying the crop.
Meanwhile the red team generated functional code actions that focused on high yield food items: sweet berries, carrots, and melons. It also had enough redundancy for both agents to keep farming carrots and melons after Byte destroyed the sweet berry bushes.
Both teams added signaling into their strategy, resulting in a lot of unnecessary waiting.

Later, in episode 17, the blue team managed to get a huge lead over the red team with 18 points. Here the tactics are quite drastically different from what they were in episode 5:
\begin{itemize}
    \item \textbf{\textcolor{red}{The red team:}} Ryn continuously harvests and replants carrots, crafts these into golden carrots and delivers 5 of these to the server, switching to beetroot soup if there are no carrots. Raze harvests and replants wheat to make bread and delivers it to the server, and then destroys all potato and sugar cane crops.
    \item \textbf{\textcolor{blue}{The blue team:}} Blink continuously harvests sweet berries and delivers them to the server. Byte harvests and replants potatoes, bakes them, and delivers 8 of them to the server.
\end{itemize}
Here the execution was notably better with all agents following the tactics correctly, and no longer waiting for signals. However red team tactics failed to counter the blue team as Blink was able to farm sweet berries uninterrupted, gaining their team 18 points, and Byte had already harvested all the potatoes it needed before Raze went and destroyed the crops.
Also the red team got quite unlucky with low carrot yields and a Raze being unable to successfully replant the wheat due to circumstance.
The red team rectified this in the next episode by having Raze destroy sweet berry bushes and craft baked potatoes instead of bread.

\textbf{Learned Causal Graphs.}
After 20 episodes of self-play, TactiCrafter learned causal graphs that are pretty accurate and model most of the dependencies the Causal Model \textit{can} model. 
We can see a lot of near identical actions with different counts like ``\texttt{mineBlock(bot, 'slime\_block', 1);}'' and ``\texttt{mineBlock(bot, 'slime\_block', 2);}''. This makes sense, because the Causal Model does not know exactly which actions the agent is executing and instead has to infer it from the chat log which only shows the exact amount of blocks actually mined which could differ from the count in the \texttt{mineBlock} argument which is an upper bound.
Further we see that cause and effects listed in the learned causal graphs only concern items present in the agent's inventory, so cause and effects are often empty. This is by design, because these concern things like block or entity states in the world which are invisible to the Causal Model in TactiCrafter.
While the learned causal graphs achieve high accuracy in capturing inventory-related dependencies, their scope remains inherently limited. Since the causal model is based solely on inventory states, it fails to represent interactions involving block states, entity behaviors, or spatial-temporal dependencies. As a result, it can only capture a fraction of the game mechanics present in the scenario. To fully model the underlying dynamics, we need more sophisticated methods that extend beyond inventory-based causal discovery to incorporate spatial reasoning, environmental states, and temporal sequences.
In the examples listed in Appendix~\ref{learned_graphs}, the listed causes for each action are 100\% accurate, and the listed effects are 94.5\% accurate. Errors in effects include: Missing buckets after cake recipe, missing wheat seeds after harvesting wheat, missing poisonous potato after harvesting potato, extraneous pumpkin seeds after harvesting pumpkins, missing feather after killing chicken, and farmland after tilling land.
Most of these errors were already in the causal graph since the first episode, so it is likely that the problematic actions got generated by the initial causal graph generation step and never got updated due to the agent never deciding to execute these actions.

\section{Conclusion}
This paper introduced PillagerBench, a benchmark for evaluating multi-agent collaboration in competitive Minecraft environments, alongside TactiCrafter, a team-based gameplay system. We explored how LLM-based agents adapt to team-vs-team scenarios, adjust to opponent strategies, and improve via self-play. Our results show that TactiCrafter outperforms random and other LLM-based baselines, demonstrating strong tactical reasoning and cooperation. These findings provide a solid foundation for future research in real-time multi-agent collaboration and competitive gameplay in complex settings.

\printbibliography

\newpage
\onecolumn

\appendix
\label{sec:appendix}

\section{Learned Causal Graphs}
\label{learned_graphs}
This section gives a few examples of the learned causal graphs by TactiCrafter playing against itself for 20 consecutive episodes in both scenarios in PillagerBench. Listing~\ref{lst:selfplay_graph_m} shows the graph for Mushroom war, and listing~\ref{lst:selfplay_graph_d} shows the graph for Dash and Dine.
\begin{lstlisting}[caption={Learned causal graph of the red team after 20 consecutive episodes of self-play in Mushroom War.},label={lst:selfplay_graph_m}]
Action: mineBlock(bot, \"slime_block\", 1); Cause: []; Effect ['slime_block']
Action: placeItem(bot, \"slime_block\", bot.entity.position.offset(1, 0, 0)); Cause: ['slime_block']; Effect []
Action: sendSignal(bot, \"Ryn\"); Cause: []; Effect []
Action: waitSignal(bot, null); Cause: []; Effect []
Action: killMob(bot, \"pig\", 300); Cause: []; Effect ['raw_porkchop']
Action: giveToPlayer(bot, \"raw_porkchop\", \"Ryn\", 1); Cause: ['raw_porkchop']; Effect []
Action: sendSignal(bot, \"Raze\"); Cause: []; Effect []
Action: mineBlock(bot, \"red_mushroom_block\", 1); Cause: []; Effect ['red_mushroom']
Action: mineBlock(bot, \"slime_block\", 8); Cause: []; Effect ['slime_block']
Action: mineBlock(bot, 'slime_block', 8); Cause: []; Effect ['slime_block']
Action: mineBlock(bot, 'red_mushroom_block', 1); Cause: []; Effect ['red_mushroom']
Action: sendSignal(bot, 'Raze'); Cause: []; Effect []
Action: mineBlock(bot, 'slime_block', 1); Cause: []; Effect ['slime_block']
Action: mineBlock(bot, 'slime_block', 10); Cause: []; Effect ['slime_block']
Action: mineBlock(bot, 'slime_block', 2); Cause: []; Effect ['slime_block']
Action: mineBlock(bot, 'slime_block', 7); Cause: []; Effect ['slime_block']
Action: mineBlock(bot, 'slime_block', 4); Cause: []; Effect ['slime_block']
Action: placeItem(bot, 'slime_block', bot.entity.position.offset(1, 0, 0)); Cause: ['slime_block']; Effect []
Action: mineBlock(bot, 'slime_block', 9); Cause: []; Effect ['slime_block']
Action: mineBlock(bot, 'slime_block', 3); Cause: []; Effect ['slime_block']
Action: mineBlock(bot, 'slime_block', 5); Cause: []; Effect ['slime_block']
Action: mineBlock(bot, 'slime_block', 11); Cause: []; Effect ['slime_block']
Action: sendSignal(bot, 'Ryn'); Cause: []; Effect []
\end{lstlisting}
\begin{lstlisting}[caption={Learned causal graph of the red team after 20 consecutive episodes of self-play in Dash and Dine.},label={lst:selfplay_graph_d}]
Action: mineBlock(bot, 'potato', 35); Cause: []; Effect ['potato']
Action: mineBlock(bot, 'carrot', 35); Cause: []; Effect ['carrot']
Action: mineBlock(bot, 'wheat', 17); Cause: []; Effect ['wheat']
Action: mineBlock(bot, 'beetroots', 13); Cause: []; Effect ['beetroot']
Action: mineBlock(bot, 'melon', 125); Cause: []; Effect ['melon_slice']
Action: mineBlock(bot, 'pumpkin', 125); Cause: []; Effect ['pumpkin']
Action: mineBlock(bot, 'sweet_berry_bush', 160); Cause: []; Effect ['sweet_berries']
Action: mineBlock(bot, 'sugar_cane', 11); Cause: []; Effect ['sugar_cane']
Action: mineBlock(bot, 'cocoa', 54); Cause: []; Effect ['cocoa_beans']
Action: mineBlock(bot, 'egg', 2); Cause: []; Effect ['egg']
Action: smeltItem(bot, 'potato', 'coal', 35); Cause: ['furnace', 'potato', 'coal']; Effect ['baked_potato']
Action: smeltItem(bot, 'beef', 'coal', 1); Cause: ['furnace', 'beef', 'coal']; Effect ['cooked_beef']
Action: smeltItem(bot, 'chicken', 'coal', 1); Cause: ['furnace', 'chicken', 'coal']; Effect ['cooked_chicken']
Action: craftItem(bot, 'bread', 5); Cause: ['wheat']; Effect ['bread']
Action: craftItem(bot, 'beetroot_soup', 2); Cause: ['beetroot', 'bowl']; Effect ['beetroot_soup']
Action: craftItem(bot, 'cookie', 5); Cause: ['wheat', 'cocoa_beans']; Effect ['cookie']
Action: craftItem(bot, 'golden_carrot', 5); Cause: ['crafting_table', 'carrot', 'gold_nugget']; Effect ['golden_carrot']
Action: craftItem(bot, 'pumpkin_pie', 5); Cause: ['pumpkin', 'sugar', 'egg']; Effect ['pumpkin_pie']
Action: craftItem(bot, 'cake', 1); Cause: ['milk_bucket', 'sugar', 'wheat', 'egg']; Effect ['cake']
Action: placeItem(bot, 'crafting_table', bot.entity.position.offset(1, 0, 0)); Cause: ['crafting_table']; Effect []
Action: giveToPlayer(bot, 'baked_potato', 'Red_Server', -1); Cause: ['baked_potato']; Effect []
Action: giveToPlayer(bot, 'golden_carrot', 'Red_Server', -1); Cause: ['golden_carrot']; Effect []
Action: giveToPlayer(bot, 'pumpkin_pie', 'Red_Server', -1); Cause: ['pumpkin_pie']; Effect []
Action: harvestWheat(bot, new Vec3(-19, -60, -1)); Cause: []; Effect ['wheat']
Action: harvestCarrots(bot, new Vec3(-19, -60, -8)); Cause: []; Effect ['carrot']
Action: harvestPotatoes(bot, new Vec3(-19, -60, 6)); Cause: []; Effect ['potato']
Action: harvestBeetroot(bot, new Vec3(-15, -60, 6)); Cause: []; Effect ['beetroot', 'beetroot_seeds']
Action: harvestMelons(bot, new Vec3(-15, -60, -1)); Cause: []; Effect ['melon_slice']
Action: harvestPumpkins(bot, new Vec3(-15, -60, -1)); Cause: []; Effect ['pumpkin', 'pumpkin_seeds']
Action: harvestSugarCane(bot, -61); Cause: []; Effect ['sugar_cane']
Action: harvestBerries(bot, new Vec3(-11, -60, 6)); Cause: []; Effect ['sweet_berries']
Action: harvestCocoa(bot, new Vec3(-15, -60, -8)); Cause: []; Effect ['cocoa_beans']
Action: milkCow(bot); Cause: ['bucket']; Effect ['milk_bucket']
Action: killMob(bot, 'cow', 300); Cause: []; Effect ['beef', 'leather']
Action: killMob(bot, 'chicken', 300); Cause: []; Effect ['chicken']
Action: getItemFromChest(bot, new Vec3(-3, -59, -1), {'bowl': 10}); Cause: []; Effect ['bowl']
Action: getItemFromChest(bot, new Vec3(-7, -60, 5), {'egg': 10}); Cause: []; Effect ['egg']
Action: getItemFromChest(bot, new Vec3(-11, -59, -1), {'gold_nugget': 64}); Cause: []; Effect ['gold_nugget']
Action: depositItemIntoChest(bot, new Vec3(-3, -59, -1), {'bowl': 10}); Cause: ['bowl']; Effect []
Action: checkItemInsideChest(bot, new Vec3(-3, -59, -1)); Cause: []; Effect []
Action: sendSignal(bot, 'Raze'); Cause: []; Effect []
Action: waitSignal(bot, null, 30000); Cause: []; Effect []
Action: tillLand(bot, ['dirt']); Cause: ['hoe']; Effect ['farmland']
Action: plantWheatSeeds(bot); Cause: ['wheat_seeds']; Effect []
Action: plantCarrots(bot); Cause: ['carrot']; Effect []
Action: plantPotatoes(bot); Cause: ['potato']; Effect []
Action: plantBeetrootSeeds(bot); Cause: ['beetroot_seeds']; Effect []
Action: plantMelonSeeds(bot); Cause: ['melon_seeds']; Effect []
Action: plantPumpkinSeeds(bot); Cause: ['pumpkin_seeds']; Effect []
Action: plantSugarCane(bot); Cause: ['sugar_cane']; Effect []
Action: plantSweetBerryBushes(bot); Cause: ['sweet_berries']; Effect []
Action: plantCocoaBeans(bot); Cause: ['cocoa_beans']; Effect []
Action: craftItem(bot, 'golden_carrot', 1); Cause: ['crafting_table', 'carrot', 'gold_nugget']; Effect ['golden_carrot']
Action: getItemFromChest(bot, new Vec3(-7, -60, 5), {'egg': 1}); Cause: []; Effect ['egg']
Action: craftItem(bot, 'pumpkin_pie', 1); Cause: ['crafting_table', 'pumpkin', 'sugar', 'egg']; Effect ['pumpkin_pie']
Action: giveToPlayer(bot, 'golden_carrot', 'Red_Server', 2); Cause: ['golden_carrot']; Effect []
Action: giveToPlayer(bot, 'pumpkin_pie', 'Red_Server', 1); Cause: ['pumpkin_pie']; Effect []
Action: giveToPlayer(bot, 'cooked_beef', 'Red_Server', 1); Cause: ['cooked_beef']; Effect []
Action: destroyCrop(bot, 'wheat'); Cause: []; Effect ['wheat', 'wheat_seeds', 'dirt']
Action: giveToPlayer(bot, 'golden_carrot', 'Red_Server', 1); Cause: ['golden_carrot']; Effect []
Action: smeltItem(bot, 'potato', 'coal', 12); Cause: ['furnace', 'potato', 'coal']; Effect ['baked_potato']
Action: craftItem(bot, 'golden_carrot', 2); Cause: ['crafting_table', 'carrot', 'gold_nugget']; Effect ['golden_carrot']
Action: giveToPlayer(bot, 'golden_carrot', 'Red_Server', 5); Cause: ['golden_carrot']; Effect []
Action: smeltItem(bot, 'potato', 'coal', 8); Cause: ['furnace', 'potato', 'coal']; Effect ['baked_potato']
Action: craftItem(bot, 'golden_carrot', 3); Cause: ['crafting_table', 'carrot', 'gold_nugget']; Effect ['golden_carrot']
Action: giveToPlayer(bot, 'golden_carrot', 'Red_Server', 3); Cause: ['golden_carrot']; Effect []
Action: smeltItem(bot, 'potato', 'coal', 14); Cause: ['furnace', 'potato', 'coal']; Effect ['baked_potato']
Action: giveToPlayer(bot, 'golden_carrot', 'Red_Server', 0); Cause: ['golden_carrot']; Effect []
Action: smeltItem(bot, 'potato', 'coal', 1); Cause: ['furnace', 'potato', 'coal']; Effect ['baked_potato']
Action: giveToPlayer(bot, 'baked_potato', 'Red_Server', 1); Cause: ['baked_potato']; Effect []
Action: smeltItem(bot, 'potato', 'coal', 2); Cause: ['furnace', 'potato', 'coal']; Effect ['baked_potato']
Action: giveToPlayer(bot, 'baked_potato', 'Red_Server', 2); Cause: ['baked_potato']; Effect []
Action: smeltItem(bot, 'potato', 'coal', 9); Cause: ['furnace', 'potato', 'coal']; Effect ['baked_potato']
Action: smeltItem(bot, 'potato', 'coal', 5); Cause: ['furnace', 'potato', 'coal']; Effect ['baked_potato']
Action: giveToPlayer(bot, 'baked_potato', 'Red_Server', 5); Cause: ['baked_potato']; Effect []
Action: smeltItem(bot, 'potato', 'coal', 4); Cause: ['furnace', 'potato', 'coal']; Effect ['baked_potato']
Action: smeltItem(bot, 'potato', 'coal', 13); Cause: ['furnace', 'potato', 'coal']; Effect ['baked_potato']
Action: smeltItem(bot, 'potato', 'coal', 3); Cause: ['furnace', 'potato', 'coal']; Effect ['baked_potato']
Action: giveToPlayer(bot, 'baked_potato', 'Red_Server', 3); Cause: ['baked_potato']; Effect []
Action: destroyCrop(bot, 'sugar_cane'); Cause: []; Effect ['sugar_cane']
Action: destroyCrop(bot, 'pumpkin'); Cause: []; Effect ['pumpkin', 'pumpkin_seeds', 'dirt']
Action: giveToPlayer(bot, 'golden_carrot', 'Red_Server', 4); Cause: ['golden_carrot']; Effect []
Action: giveToPlayer(bot, 'baked_potato', 'Red_Server', 8); Cause: ['baked_potato']; Effect []
Action: smeltItem(bot, 'potato', 'coal', 10); Cause: ['furnace', 'potato', 'coal']; Effect ['baked_potato']
Action: smeltItem(bot, 'potato', 'coal', 20); Cause: ['furnace', 'potato', 'coal']; Effect ['baked_potato']
Action: craftItem(bot, 'beetroot_soup', 1); Cause: ['crafting_table', 'beetroot', 'bowl']; Effect ['beetroot_soup']
Action: giveToPlayer(bot, 'beetroot_soup', 'Red_Server', 1); Cause: ['beetroot_soup']; Effect []
\end{lstlisting}

\end{document}